\documentclass{article}

\usepackage{microtype}
\usepackage{graphicx}
\usepackage{subcaption}
\usepackage{booktabs} 
\usepackage{float}
\usepackage{adjustbox}
\usepackage{multirow}
\usepackage{hyperref}

\usepackage[accepted]{icml2026}

\usepackage{amsmath}
\usepackage{amssymb}
\usepackage{mathtools}
\usepackage{amsthm}

\usepackage[capitalize,noabbrev]{cleveref}

\theoremstyle{plain}

\theoremstyle{definition}

\theoremstyle{remark}

\icmltitlerunning{Memory Efficient Tabular Foundation Models}

\begin{document}

\twocolumn[
  \icmltitle{Memory Efficient Tabular Foundation Models}



  \icmlsetsymbol{equal}{*}

\begin{icmlauthorlist}
  \icmlauthor{Shuting Luo}{lab}
  \icmlauthor{Monika Mikhail Kanaan}{lab}
  \icmlauthor{Cameron Gordon}{adl}
  \icmlauthor{Anna Leontjeva}{lab}
  \icmlauthor{Simon Lucey}{adl}
\end{icmlauthorlist}

\icmlaffiliation{lab}{Commonwealth Bank of Australia, Australia}

\icmlaffiliation{adl}{Australian Institute for Machine Learning, University of Adelaide, Adelaide, Australia}

\icmlcorrespondingauthor{Shuting Luo}{aria.luo@cba.com.au}

  \icmlkeywords{Machine Learning, ICML}

  \vskip 0.3in
]



\printAffiliationsAndNotice{}  

\begin{abstract}
  Tabular Foundation Models, such as TabPFN, have received a large amount of recent attention due to their performance on in-context tabular machine learning tasks, which often exceeds classical baselines. However, practical deployment considerations of these models have received less attention. In this paper we investigate the memory requirements for these models. We demonstrate that employing model compression approaches can enable memory reductions of up to 7.6$\times$ with similar levels of performance, reducing deployment requirements by nearly $87\%$. Our work provides insight to practitioners seeking efficient deployment of these models in practical settings. 
  
\end{abstract}

\section{Introduction} 













Tabular machine learning is used heavily in practical domains such as healthcare, finance, operations, and enterprise analytics, where models are often deployed under strict constraints on latency, cost, hardware availability, and data movement~\citep{jiangrepresentation,shwartz-ziv2021tabular,borisov2024}. Recent tabular foundation models (TFMs), including the TabPFN and TabICL series, are attractive in these settings because they can make strong predictions in-context, reducing the need for task-specific training, hyperparameter tuning, and bespoke model-selection pipelines~\citep{jiangrepresentation, hollmann2023tabpfn,qu2025tabicl, TabPFN-2.5}. This makes them appealing for organizations that need to solve many small or medium-sized tabular prediction problems repeatedly.

However, practical adoption depends not only on predictive performance, but also on whether these models can be deployed and communicated efficiently~\citep{fu2024serverlessllm}. Enterprises must contend with memory movements, hardware constraints, and practical bandwidth considerations. These issues are well-known constraints in servicing large language models, where memory footprint directly affects deployment cost, latency, and the feasibility of local inference~\citep{Menghani2023,Jeyaraman2025}. To address this, memory-efficient techniques such as model quantization are now ubiquitous in practical pipelines~\citep{dettmers2023qlora,gholami2022survey}. 

These memory-based limitations also exist for TFMs, however to date most work has focused on reducing \textit{context} constraints, due to architectural limitations and attention-based inference that grows quickly with sample and feature dimensions~\citep{zabergja2026end,TabPFN-2.5,qu2026tabiclv2betterfasterscalable,zeng2024tabflex,liutabpfn}, leaving practical deployment considerations relatively unexplored.

\begin{figure}
    \centering
    \includegraphics[width=\linewidth]{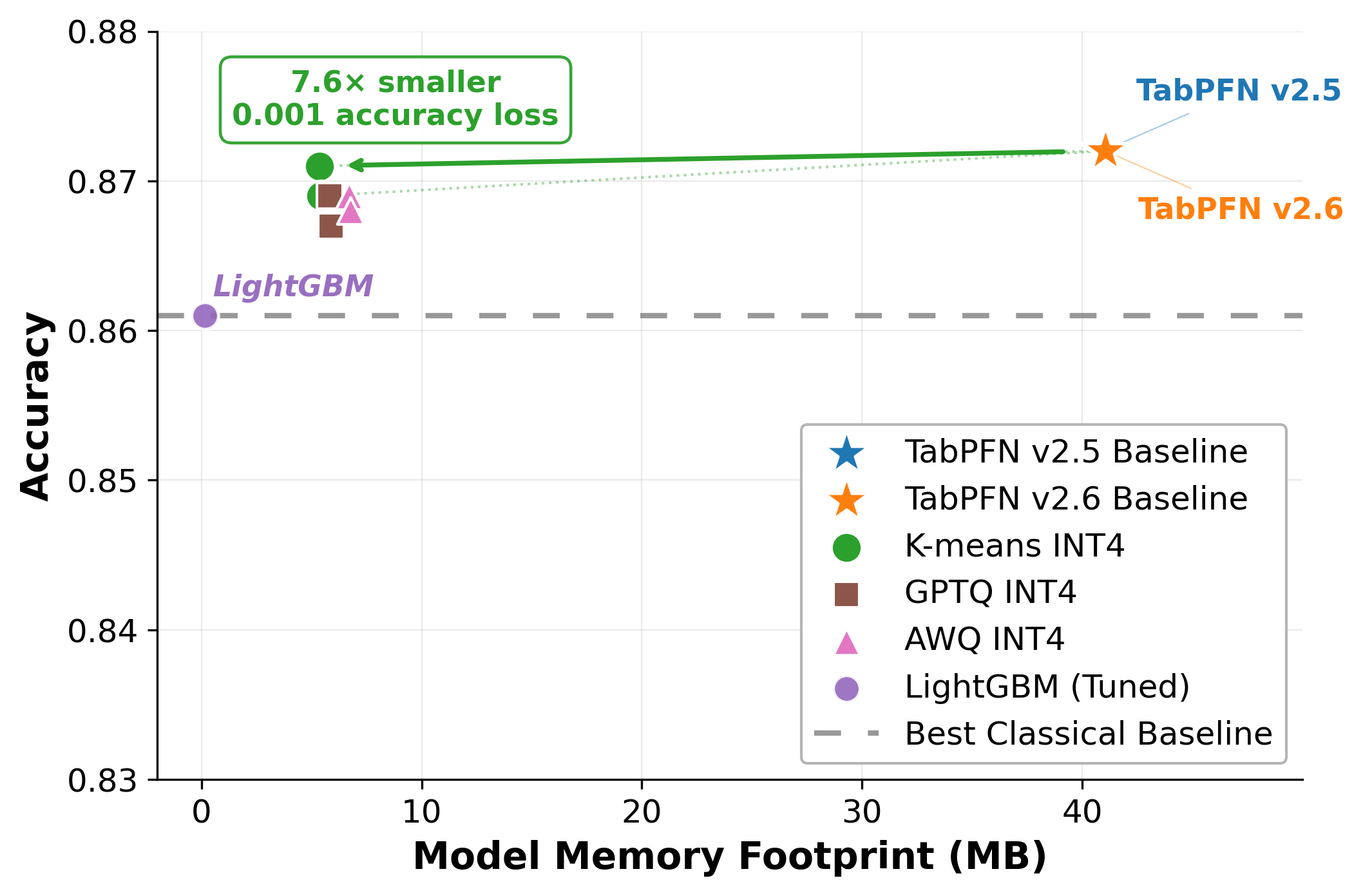}
    \caption{INT4 quantization makes Tabular Foundation Models substantially more memory-efficient while preserving their accuracy advantage. Quantized models have a $7.6\times$ lower memory footprint relative to full precision baselines, with negligible decline in accuracy, remaining above the strongest tuned classical baseline. This demonstrates the potential for quantized TFMs to achieve high predictive performance at a fraction of the deployment cost.}
    \label{fig:front-figure}
\end{figure}

This paper asks the question: \textbf{Can we reduce the deployment requirements for TFMs while retaining their predictive performance?} We evaluate whether existing quantization methods can be employed to reduce the memory footprint of the TabPFN family of models, using standard tabular benchmarks and classical baselines for context.


\begin{figure*}[htbp]
    \centering

    \begin{subfigure}[b]{0.48\textwidth}
        \centering
        \includegraphics[width=\textwidth]{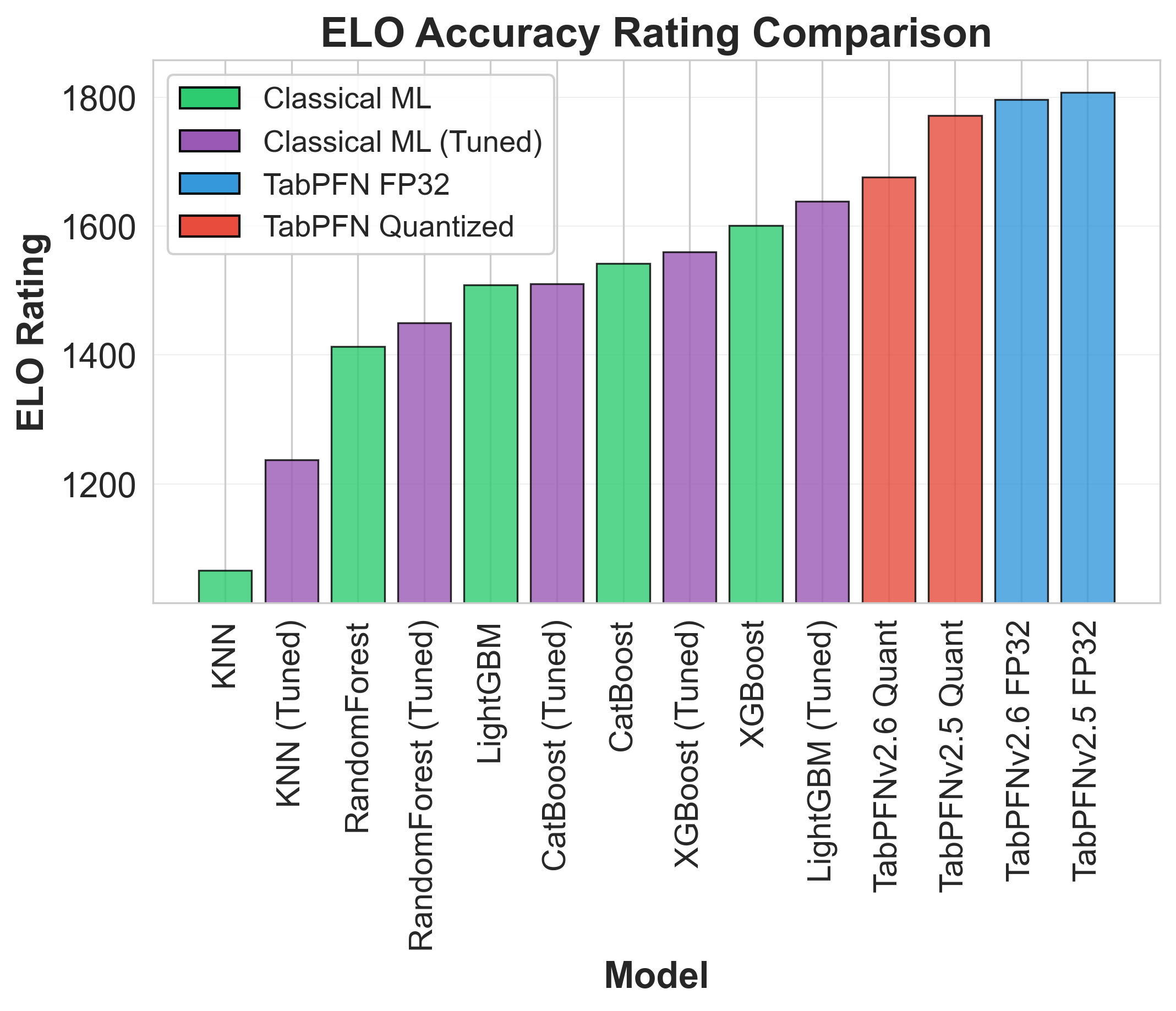}
        \caption{ELO ratings from pairwise accuracy comparisons. Quantized TabPFN models remain close to their FP32 counterparts and rank above all tuned and untuned classical baselines.}
        \label{fig:elo_accuracy}
    \end{subfigure}
    \hfill
    \begin{subfigure}[b]{0.48\textwidth}
        \centering
        \includegraphics[width=\textwidth]{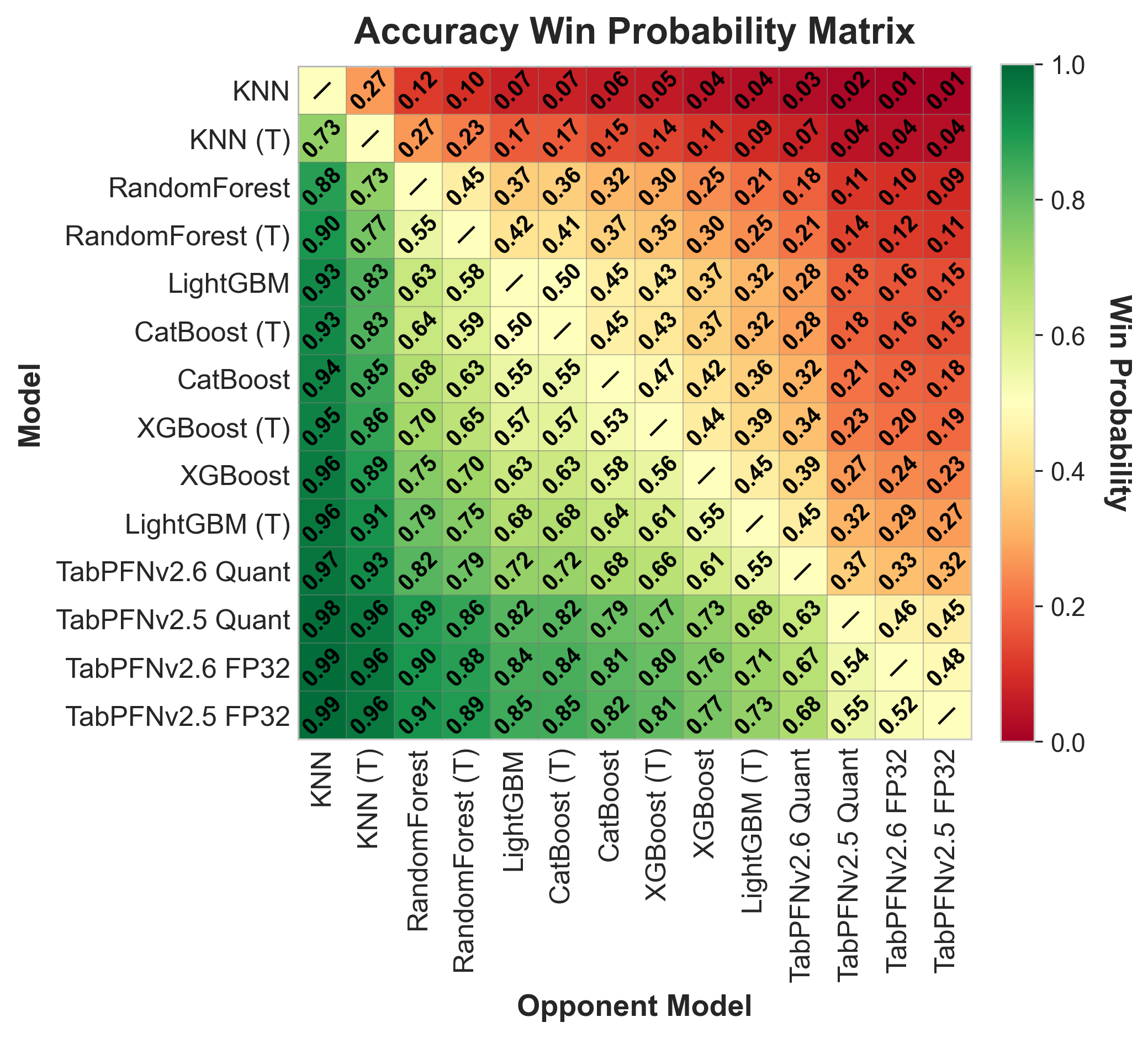}
        \caption{Pairwise win probabilities on evaluated datasets. Quantized TabPFN models consistently beat classical baselines, with most losses concentrated against FP32 TabPFN variants.}
        \label{fig:accuracy_win_matrix}
    \end{subfigure}

    \caption{Quantization preserves most of the accuracy gains of Tabular Foundation Models. Quantized TabPFN variants incur only a modest drop relative to FP32 models while maintaining a clear advantage over strong classical ML baselines.}
    \label{fig:elo_winprob_accuracy}
\end{figure*}

Our results indicate that quantization is a simple and practical route to memory-efficient TFMs. Across 30 datasets, we find that quantizing to INT4 precision reduces memory footprint of TFMs while maintaining similar predictive performance to the full precision counterparts. These results show that deployment needs can be reduced by up to 7.6$\times$ while maintaining performance comparable to tree-based models like CatBoost~\citep{prokhorenkova2018catboost}.


In this paper we make the following contributions: 
\begin{itemize}
    \item We highlight the practical need for memory-efficient tabular machine learning models and demonstrate that large improvements can be obtained using standard quantization methods. 
    \item We provide a systematic evaluation of these quantization approaches, and qualitative considerations for practical deployment.  
\end{itemize}

We demonstrate our experiments on a range of standard tabular datasets using the TabPFNv2.5 and TabPFNv2.6 variants from the TabPFN family of models~\citep{TabPFN-2.5}.

\begin{table*}[t]
\centering
\caption{Comparison of model performance and compression methods (30 Datasets).}
\label{tab:model_compression_comparison30}
\begin{tabular}{llccccc}
\hline
\textbf{Model} & \textbf{Method} & \textbf{Accuracy} & \textbf{AUC} & \textbf{Memory (MB)} & \textbf{Compression Ratio} \\
\hline

\multirow{4}{*}{TabPFN-v2.5}
& Baseline    & 0.872 & 0.869 & 40.95 & 1.00x \\
& GPTQ INT4   & 0.869 & 0.866 & 5.81  & 7.04x \\
& KMeans INT4 & 0.871 & 0.867 & 5.36  & 7.65x \\
& AWQ INT4    & 0.869 & 0.866 & 6.70  & 6.12x \\
\hline

\multirow{4}{*}{TabPFN-v2.6}
& Baseline    & 0.872 & 0.870 & 41.05 & 1.00x \\
& GPTQ INT4   & 0.867 & 0.865 & 5.89  & 6.97x \\
& KMeans INT4 & 0.869 & 0.868 & 5.43  & 7.56x \\
& AWQ INT4    & 0.868 & 0.866 & 6.77  & 6.06x \\
\hline

\multirow{2}{*}{XGBoost}
& Baseline & 0.860 & 0.850 & 0.096 & -- \\
& Tuned    & 0.862 & 0.855 & 0.142 & -- \\
\hline

\multirow{2}{*}{CatBoost}
& Baseline & 0.858 & 0.851 & 0.073 & -- \\
& Tuned    & 0.861 & 0.856 & 0.099 & -- \\
\hline

\multirow{2}{*}{LightGBM}
& Baseline & 0.860 & 0.851 & 0.124 & -- \\
& Tuned    & 0.864 & 0.854 & 0.183 & -- \\
\hline

\multirow{2}{*}{KNN}
& Baseline & 0.798 & 0.704 & 0.203 & -- \\
& Tuned    & 0.828 & 0.755 & 0.220 & -- \\
\hline

\multirow{2}{*}{RandomForest}
& Baseline & 0.859 & 0.850 & 0.367 & -- \\
& Tuned    & 0.861 & 0.852 & 0.853 & -- \\
\hline

\end{tabular}
\end{table*}

\section{Related Literature} 
\subsection{Tabular Foundation Models}

Tabular Foundation Models are a class of pre-trained models that can be applied to tabular machine learning problems without additional training~\citep{jiangrepresentation}. These models are trained to predict a posterior prediction ($y_{test})$ based on a tuple of in-context data ($x_{train}, y_{train}; x_{test}$)~\citep{hollmann2023tabpfn,qu2025tabicl}. The training of these models involves extensive pre-training on simulated data of causal relationships and are known as Prior Fitted Networks (PFNs)~\citep{muller2022transformers, hollmann2023tabpfn}. Follow up work has sought to extend these models to more realistic tabular scenarios (e.g. missing data, categorical data, time-series), and to handle limitations in the feature and sample dimension~\citep{hollmann2025tabpfn, TabPFN-2.5, qu2025tabicl,qu2026tabiclv2betterfasterscalable, hoo2025tables}. The strong performance of these models relative to classical baselines has generated applied interest in a wide range of practical domains such as medical machine learning, financial forecasting, and a variety of industrial uses~\citep{TabPFN-2.5}. The practical handling of large contexts has received a large amount of interest~\citep{zeng2024tabflex, zabergja2026end}. However, memory requirements of weight deployment have received relatively little attention. 

\subsection{Quantization and Model Compression} 
Quantization is a mapping involving the discretization of a signal~\citep{gersho1991vector}. Within machine learning, the term is typically used to refer to \textit{weight} or \textit{activation} quantization, where the goal is to reduce memory by using lower precision values~\citep{gholami2022survey}.\footnote{Note: the term \textit{quantization} is occasionally used to refer to discretization of features in tabular machine learning contexts. We use the term to refer to weight quantization in this paper reflecting our interest in memory efficiency in Tabular Foundation Models.}
 We outline briefly broad trends in the literature and key design decisions. 

Early work includes binary neural networks~\citep{rastegari_2016, NIPS2016_d8330f85}, the use of quantization aware-training, and integer quantization~\citep{jacob2018quantization}. Classical results show that MSE optimal quantizers correspond closely to a clustering problem~\citep{gersho1991vector}, motivating codebook k-means quantizers~\citep{han2016deepcompressioncompressingdeep}. 

Recent advances in quantization have included the use of random rotations~\citep{zandieh2026turboquant, liuspinquant, ashkboos2024quarot}; the use of singular value decomposition to reduce the effect of outlier values~\citep{li2024svdquant}; activation-aware quantization (AWQ)~\citep{lin2023awq}; and dedicated libraries such as GPTQ and BitsandBytes~\citep{dettmers2022llmint8,dettmers2022optimizers,dettmers2023qlora, frantar-gptq}. A large body of literature exists on applying quantization methods to reduce the memory requirements for modality specific use-cases such as large language model compression~\citep{gholami2022survey}. However, little work has been devoted to weight quantization within TFMs.

\subsection{Memory Efficient Tabular Models}
There is some work in reducing memory requirements for tabular machine learning models. For example, through the use of low-precision decision gradient boosted trees~\citep{herrmann2025boosted,shi2022quantized}; and in applying integer quantization to enable efficient performance on FPGA-enabled edge devices~\citep{Alsharari2025}. Separately, within TFMs there are some works on achieving high tabular performance with small MLP or distilled models such as TabM~\citep{gorishniy2025tabm} and TabDistill~\citep{dissanayake2025tabdistill}. In addition, inference memory requirements have been reduced by compressing context~\citep{zabergja2026end}, or through hypernetworks~\citep{muellermothernet}. 

\section{Method} 





\subsection{Quantization Methods}

We evaluate three INT4 post-training quantization approaches 
indicative of recent trends in model compression: K-Means 
Quantization~\citep{han2016deepcompressioncompressingdeep}, 
AWQ~\citep{lin2023awq}, and GPTQ~\citep{frantar-gptq}. We focus 
on post-training quantization (PTQ) rather than 
quantization-aware training, as this can be conducted on a pre-trained model and does not require access 
to the original pipeline. For GPTQ and 
AWQ, originally developed for autoregressive language models, 
we adapt their calibration procedures to TabPFN's in-context 
learning setting by treating the $(x_{train}, y_{train}; 
x_{test})$ tuple as the calibration input. For K-Means 
clustering quantization, we implement an INT4 bit-packing 
scheme to reduce model artifact size on disk.

\begin{figure}
    \centering
    \includegraphics[width=\linewidth]{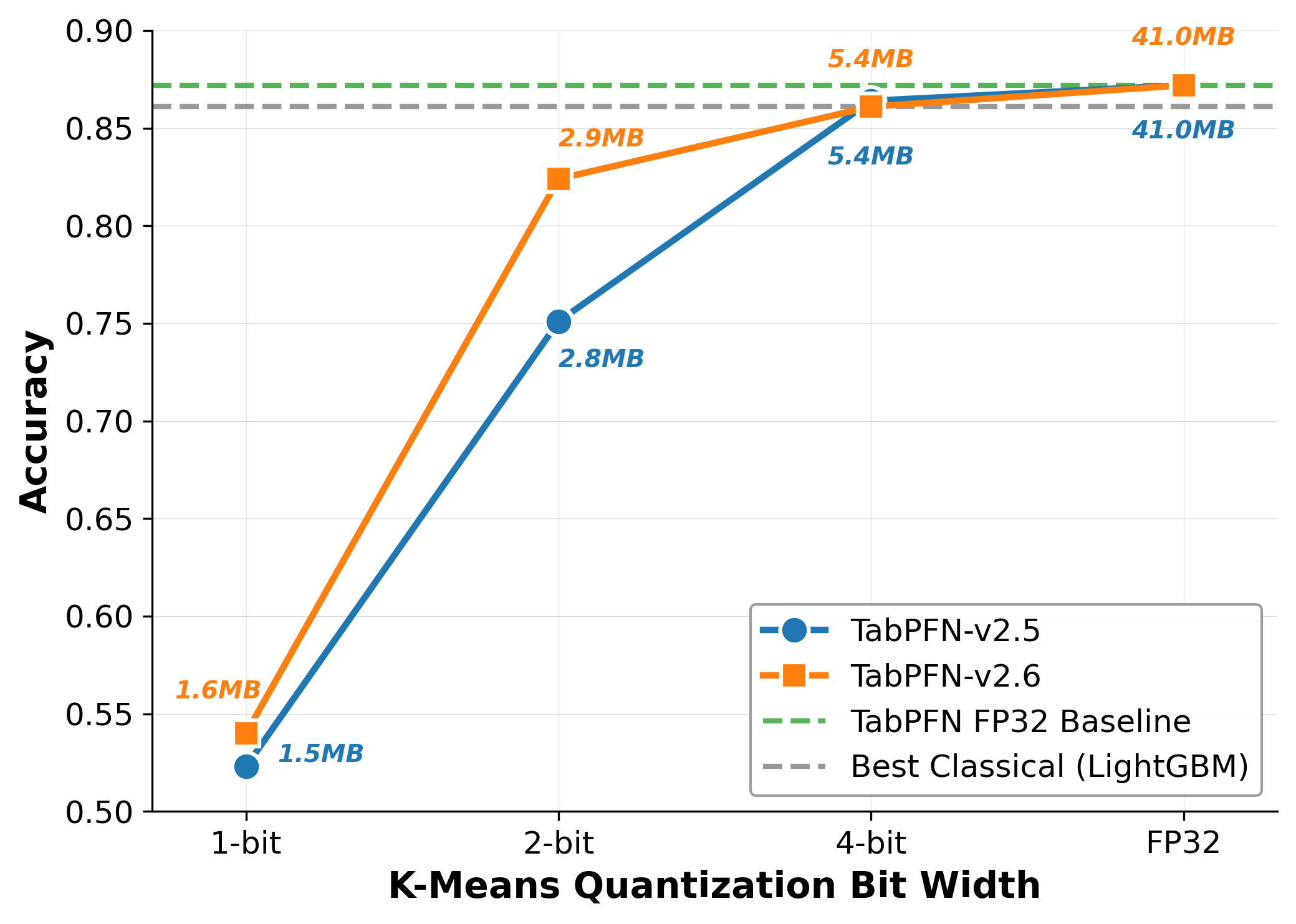}
    \caption{Effect of K-means quantization bit width on TabPFN accuracy and model size. Accuracy improves sharply from 1-bit to 2-bit quantization and largely saturates by 4-bit quantization, where both TabPFN-v2.5 and TabPFN-v2.6 match or approach full precision performance while requiring only 5.4 MB. Further increasing to FP32 yields negligible accuracy gains but increases model size substantially to 41.0 MB, indicating that 4-bit quantization provides the best accuracy-memory trade-off.}
    \label{fig:placeholder}
\end{figure}

\subsection{Experimental Setup}

We evaluate on a subset of 30 datasets from the OpenML 
library~\citep{Vanschoren2013}, selected to match feature 
and sample limitations of TabPFN. A full list of datasets 
is provided in Appendix~\ref{sec:datasets}. We test our three 
quantization methods on two TabPFN 
variants:  
TabPFNv2.5, and 
TabPFNv2.6~\citep{TabPFN-2.5}. Memory is evaluated following BZIP2 compression, replicating a practical deployment pipeline~\citep{seward_bzip2_1996}. We benchmark against standard 
tabular machine learning methods: XGBoost~\citep{chen2016xgboost}, 
CatBoost~\citep{prokhorenkova2018catboost}, 
LightGBM~\citep{ke2017lightgbm}, 
KNN~\citep{fix1951discriminatory, cover1967nearest}, 
and Random Forest~\citep{breiman2001random}, evaluated in 
both default-configuration and tuned variants. Hyperparameter details are provided in Appendix ~\ref{sec:baselines}.

\section{Results}


Table~\ref{tab:model_compression_comparison30} and Figure~\ref{fig:front-figure} present 
quantization performance across TabPFNv2.5 and TabPFNv2.6. All three INT4 quantization methods achieve 
substantial compression (6.0-7.65$\times$) with minimal accuracy degradation 
($\leq0.005$). Among quantization methods, K-means INT4 achieves the highest compression 
ratio (7.65$\times$ for TabPFNv2.5, and 7.56$\times$ for TabPFNv2.6) while incurring 
the smallest accuracy loss (0.001), outperforming both GPTQ and AWQ 
across both model versions. Critically, all quantized variants 
maintain accuracy above the best classical baseline (LightGBM tuned: 0.864), demonstrating that compression preserves TabPFN's performance 
advantage over traditional methods.

Several further observations support these results. First, a 
bit-width ablation (Table~\ref{tab:kmeans_nbit_ablation}) confirms 
INT4 as the practical optimum: 2-bit quantization recovers only 
$\sim$90\% of baseline accuracy and 1-bit collapses to near-random 
performance, while INT4 retains over 99\%. Second, Figure~\ref{fig:elo_winprob_accuracy} shows that the quantized models retain a consistent dataset-level advantage over classical baselines, with a marginal relative ELO loss relative to full precision versions. Third, performance variance across the 
30 datasets remains essentially unchanged between baseline and 
quantized variants (see Appendix Table~\ref{tab:stability}), indicating that 
quantization does not introduce dataset-dependent instability. 
Finally, while deployable memory is substantially reduced, 
inference-time memory and runtime are largely unchanged 
(see Appendix Table~\ref{tab:inference_memory_speed}). This is consistent with 
TabPFN's inference cost being dominated by attention activations 
rather than weights, motivating activation-level compression as 
future work.

\section{Conclusion and Limitations}

In this work, we investigated the memory efficiency of TFMs, with a particular focus on quantization-based model compression. Our results demonstrate that substantial reductions in deployment memory are achievable with quantization without compromising predictive performance, making these models more practical for use within industry. 

\paragraph{Limitations} Our study focuses exclusively on post-training quantization, which may be suboptimal relative to quantization-aware training, which may produce higher performance at lower-precision with access to the training pipeline~\citep{rastegari_2016,han2016deepcompressioncompressingdeep,gholami2022survey}. In addition, our evaluation considers only a subset of available quantization techniques, leaving more advanced approaches such as TurboQuant~\citep{zandieh2026turboquant} and other emerging methods unexplored. As is standard in quantization for memory deployment, our method requires dequantization prior to inference and so does not enable run-time memory improvements~\citep{gholami2022survey}. We also do not incorporate kernel-level or hardware-aware optimizations, which may enable FPGA/edge-device implementations. Finally, our method retains the architectural limitations of the foundation models tested which constrain the maximum context window~\citep{TabPFN-2.5}.

\paragraph{Future Work} Several promising directions exist for extending this work. Beyond quantization, complementary forms of model compression including pruning~\citep{han2016deepcompressioncompressingdeep}, low-rank adaptation (LoRA)~\citep{hu2022lora}, and knowledge distillation~\citep{hinton2015distilling}, may yield further gains in memory and computational efficiency. Improving inference-time memory efficiency represents another valuable avenue, for example by combining memory-efficient attention mechanisms or context-handling strategies with a quantized approach. Exploring quantization-aware training may further improve efficiency while preserving low-precision predictive performance~\cite{rastegari_2016,han2016deepcompressioncompressingdeep}. Activation quantization strategies, which enable inference in low-precision are a further avenue~\citep{gholami2022survey}. Finally, integrating hardware-aware optimizations and extending TFM models to physical edge devices remains an exciting potential future direction for a range of practical memory-bound applications. 

\clearpage

\bibliography{bibliography}

@misc{TabPFN-2.5,
  title={TabPFN-2.5: Advancing the State of the Art in Tabular Foundation Models},
  author={Grinsztajn, L{\'e}o and others},
  year={2025},
  eprint={2511.08667},
  archivePrefix={arXiv},
  primaryClass={cs.LG},
  doi={10.48550/arXiv.2511.08667}
}

@inproceedings{zandieh2026turboquant,
  title={TurboQuant: Online Vector Quantization with Near-optimal Distortion Rate},
  author={Zandieh, Amir and Daliri, Majid and Hadian, Majid and Mirrokni, Vahab},
  booktitle={International Conference on Learning Representations (ICLR)},
  year={2026},
  note={Preprint available at arXiv:2504.19874}
}

@inproceedings{
qu2025tabicl,
title={Tab{ICL}: A Tabular Foundation Model for In-Context Learning on Large Data},
author={Jingang Qu and David Holzm{\"u}ller and Ga{\"e}l Varoquaux and Marine Le Morvan},
booktitle={Forty-second International Conference on Machine Learning},
year={2025},
url={https://openreview.net/forum?id=0VvD1PmNzM}
}

@misc{qu2026tabiclv2betterfasterscalable,
      title={TabICLv2: A better, faster, scalable, and open tabular foundation model}, 
      author={Jingang Qu and David Holzmüller and Gaël Varoquaux and Marine Le Morvan},
      year={2026},
      eprint={2602.11139},
      archivePrefix={arXiv},
      primaryClass={cs.LG},
      url={https://arxiv.org/abs/2602.11139}, 
}

@article{hollmann2025tabpfn,
 title={Accurate predictions on small data with a tabular foundation model},
 author={Hollmann, Noah and M{\"u}ller, Samuel and Purucker, Lennart and
         Krishnakumar, Arjun and K{\"o}rfer, Max and Hoo, Shi Bin and
         Schirrmeister, Robin Tibor and Hutter, Frank},
 journal={Nature},
 year={2025},
 month={01},
 day={09},
 doi={10.1038/s41586-024-08328-6},
 publisher={Springer Nature},
 url={https://www.nature.com/articles/s41586-024-08328-6},
}

@inproceedings{hollmann2023tabpfn,
  title={TabPFN: A transformer that solves small tabular classification problems in a second},
  author={Hollmann, Noah and M{\"u}ller, Samuel and Eggensperger, Katharina and Hutter, Frank},
  booktitle={International Conference on Learning Representations 2023},
  year={2023}
}

@book{gersho1991vector,
  title={Vector Quantization and Signal Compression},
  author={Gersho, Allen and Gray, Robert M},
  volume={159},
  year={1991},
  publisher={Springer Science \& Business Media}
}

@misc{han2016deepcompressioncompressingdeep,
      title={Deep Compression: Compressing Deep Neural Networks with Pruning, Trained Quantization and Huffman Coding}, 
      author={Song Han and Huizi Mao and William J. Dally},
      year={2016},
      eprint={1510.00149},
      archivePrefix={arXiv},
      primaryClass={cs.CV},
      url={https://arxiv.org/abs/1510.00149}, 
}

@InProceedings{rastegari_2016,
author="Rastegari, Mohammad
and Ordonez, Vicente
and Redmon, Joseph
and Farhadi, Ali",
editor="Leibe, Bastian
and Matas, Jiri
and Sebe, Nicu
and Welling, Max",
title="XNOR-Net: ImageNet Classification Using Binary Convolutional Neural Networks",
booktitle="Computer Vision -- ECCV 2016",
year="2016",
publisher="Springer International Publishing",
address="Cham",
pages="525--542",
isbn="978-3-319-46493-0"
}

@inproceedings{lin2023awq,
  title={AWQ: Activation-aware Weight Quantization for LLM Compression and Acceleration},
  author={Lin, Ji and Tang, Jiaming and Tang, Haotian and Yang, Shang and Chen, Wei-Ming and Wang, Wei-Chen and Xiao, Guangxuan and Dang, Xingyu and Gan, Chuang and Han, Song},
  booktitle={MLSys},
  year={2024}
}

@inproceedings{liuspinquant,
  title={SpinQuant: LLM Quantization with Learned Rotations},
  author={Liu, Zechun and Zhao, Changsheng and Fedorov, Igor and Soran, Bilge and Choudhary, Dhruv and Krishnamoorthi, Raghuraman and Chandra, Vikas and Tian, Yuandong and Blankevoort, Tijmen},
  booktitle={The Thirteenth International Conference on Learning Representations},
  year={2025}
}

@inproceedings{
  li2024svdquant,
  title={SVDQuant: Absorbing Outliers by Low-Rank Components for 4-Bit Diffusion Models},
  author={Li*, Muyang and Lin*, Yujun and Zhang*, Zhekai and Cai, Tianle and Li, Xiuyu and Guo, Junxian and Xie, Enze and Meng, Chenlin and Zhu, Jun-Yan and Han, Song},
  booktitle={The Thirteenth International Conference on Learning Representations},
  year={2025}
}

@inproceedings{
ashkboos2024quarot,
title={QuaRot: Outlier-Free 4-Bit Inference in Rotated {LLM}s},
author={Saleh Ashkboos and Amirkeivan Mohtashami and Maximilian L. Croci and Bo Li and Pashmina Cameron and Martin Jaggi and Dan Alistarh and Torsten Hoefler and James Hensman},
booktitle={The Thirty-eighth Annual Conference on Neural Information Processing Systems},
year={2024},
url={https://openreview.net/forum?id=dfqsW38v1X}
}

@inproceedings{NIPS2016_d8330f85,
 author = {Hubara, Itay and Courbariaux, Matthieu and Soudry, Daniel and El-Yaniv, Ran and Bengio, Yoshua},
 booktitle = {Advances in Neural Information Processing Systems},
 editor = {D. Lee and M. Sugiyama and U. Luxburg and I. Guyon and R. Garnett},
 pages = {},
 publisher = {Curran Associates, Inc.},
 title = {Binarized Neural Networks},
 url = {https://proceedings.neurips.cc/paper_files/paper/2016/file/d8330f857a17c53d217014ee776bfd50-Paper.pdf},
 volume = {29},
 year = {2016}
}

@incollection{gholami2022survey,
  title={A survey of quantization methods for efficient neural network inference},
  author={Gholami, Amir and Kim, Sehoon and Dong, Zhen and Yao, Zhewei and Mahoney, Michael W and Keutzer, Kurt},
  booktitle={Low-power computer vision},
  pages={291--326},
  year={2022},
  publisher={Chapman and Hall/CRC}
}

@article{dettmers2023qlora,
  title={Qlora: Efficient finetuning of quantized llms},
  author={Dettmers, Tim and Pagnoni, Artidoro and Holtzman, Ari and Zettlemoyer, Luke},
  journal={arXiv preprint arXiv:2305.14314},
  year={2023}
}

@article{dettmers2022optimizers,
  title={8-bit Optimizers via Block-wise Quantization},
  author={Dettmers, Tim and Lewis, Mike and Shleifer, Sam and Zettlemoyer, Luke},
  journal={9th International Conference on Learning Representations, ICLR},
  year={2022}
}

@article{dettmers2022llmint8,
  title={LLM.int8(): 8-bit Matrix Multiplication for Transformers at Scale},
  author={Dettmers, Tim and Lewis, Mike and Belkada, Younes and Zettlemoyer, Luke},
  journal={arXiv preprint arXiv:2208.07339},
  year={2022}
}

@article{frantar-gptq,
  title={{GPTQ}: Accurate Post-training Compression for Generative Pretrained Transformers}, 
  author={Elias Frantar and Saleh Ashkboos and Torsten Hoefler and Dan Alistarh},
  year={2022},
  journal={arXiv preprint arXiv:2210.17323}
}

@inproceedings{jacob2018quantization,
  title={Quantization and training of neural networks for efficient integer-arithmetic-only inference},
  author={Jacob, Benoit and Kligys, Skirmantas and Chen, Bo and Zhu, Menglong and Tang, Matthew and Howard, Andrew and Adam, Hartwig and Kalenichenko, Dmitry},
  booktitle={Proceedings of the IEEE conference on computer vision and pattern recognition},
  pages={2704--2713},
  year={2018}
}

@inproceedings{chen2016xgboost,
  title={Xgboost: A scalable tree boosting system},
  author={Chen, Tianqi and Guestrin, Carlos},
  booktitle={Proceedings of the 22nd acm sigkdd international conference on knowledge discovery and data mining},
  pages={785--794},
  year={2016}
}

@article{ke2017lightgbm,
  title={Lightgbm: A highly efficient gradient boosting decision tree},
  author={Ke, Guolin and Meng, Qi and Finley, Thomas and Wang, Taifeng and Chen, Wei and Ma, Weidong and Ye, Qiwei and Liu, Tie-Yan},
  journal={Advances in neural information processing systems},
  volume={30},
  year={2017}
}

@article{prokhorenkova2018catboost,
  title={CatBoost: unbiased boosting with categorical features},
  author={Prokhorenkova, Liudmila and Gusev, Gleb and Vorobev, Aleksandr and Dorogush, Anna Veronika and Gulin, Andrey},
  journal={Advances in neural information processing systems},
  volume={31},
  year={2018}
}

@article{jiangrepresentation,
  title={Representation learning for tabular data: A comprehensive survey},
  author={Jiang, Jun-Peng and Liu, Si-Yang and Cai, Hao-Run and Zhou, Qi-Le and Ye, Han-Jia},
  journal={IEEE Transactions on Pattern Analysis and Machine Intelligence},
  year={2026},
  publisher={IEEE}
}

@inproceedings{
muller2022transformers,
title={Transformers Can Do Bayesian Inference},
author={Samuel M{\"u}ller and Noah Hollmann and Sebastian Pineda Arango and Josif Grabocka and Frank Hutter},
booktitle={International Conference on Learning Representations},
year={2022},
url={https://openreview.net/forum?id=KSugKcbNf9}
}

@article{hoo2025tables,
  title={From Tables to Time: Extending TabPFN-v2 to Time Series Forecasting},
  author={Hoo, Shi Bin and M{\"u}ller, Samuel and Salinas, David and Hutter, Frank},
  journal={arXiv preprint arXiv:2501.02945},
  year={2025}
}

@article{shi2022quantized,
  title={Quantized training of gradient boosting decision trees},
  author={Shi, Yu and Ke, Guolin and Chen, Zhuoming and Zheng, Shuxin and Liu, Tie-Yan},
  journal={Advances in neural information processing systems},
  volume={35},
  pages={18822--18833},
  year={2022}
}

@ARTICLE{Alsharari2025,
  author={Alsharari, Majed and Mai, Son T. and Woods, Roger and Reaño, Carlos},
  journal={IEEE Transactions on Circuits and Systems I: Regular Papers}, 
  title={Efficient Integer-Only-Inference of Gradient Boosting Decision Trees on Low-Power Devices}, 
  year={2025},
  volume={72},
  number={1},
  pages={241-253},
  doi={10.1109/TCSI.2024.3446582}}

@article{herrmann2025boosted,
  title={Boosted Trees on a Diet: Compact Models for Resource-Constrained Devices},
  author={Herrmann, Nina and Stenkamp, Jan and Karic, Benjamin and Oehmcke, Stefan and Gieseke, Fabian},
  journal={arXiv preprint arXiv:2510.26557},
  year={2025}
}

@inproceedings{
gorishniy2025tabm,
title={TabM: Advancing tabular deep learning with parameter-efficient ensembling},
author={Yury Gorishniy and Akim Kotelnikov and Artem Babenko},
booktitle={The Thirteenth International Conference on Learning Representations},
year={2025},
url={https://openreview.net/forum?id=Sd4wYYOhmY}
}

@inproceedings{muellermothernet,
  title={MotherNet: Fast Training and Inference via Hyper-Network Transformers},
  author={Mueller, Andreas C. and Curino, Carlo and Ramakrishnan, Raghu},
  booktitle={International Conference on Learning Representations},
  year={2025}
}

@article{dissanayake2025tabdistill,
  title={TabDistill: Distilling Transformers into Neural Nets for Few-Shot Tabular Classification},
  author={Dissanayake, Pasan and Dutta, Sanghamitra},
  journal={arXiv preprint arXiv:2511.05704},
  year={2025}
}

@article{Vanschoren2013,
  author = {Joaquin Vanschoren and Jan N. van Rijn and Bernd Bischl and Luis Torgo},
  title = {OpenML: Networked Science in Machine Learning},
  journal = {SIGKDD Explorations},
  volume = {15},
  number = {2},
  pages = {49--60},
  year = {2013},
  doi = {10.1145/2641190.2641198},
  url = {https://doi.org/10.1145/2641190.2641198}
}

@article{breiman2001random,
  title={Random forests},
  author={Breiman, Leo},
  journal={Machine Learning},
  volume={45},
  number={1},
  pages={5--32},
  year={2001},
  publisher={Springer},
  doi={10.1023/A:1010933404324}
}

@techreport{fix1951discriminatory,
  title={Discriminatory analysis. nonparametric discrimination: small sample performance},
  author={Fix, Evelyn and Hodges Jr, Joseph L},
  year={1951},
  institution={California Univ Berkeley}
}

@article{cover1967nearest,
  title={Nearest neighbor pattern classification},
  author={Cover, Thomas and Hart, Peter},
  journal={IEEE transactions on information theory},
  volume={13},
  number={1},
  pages={21--27},
  year={1967},
  publisher={IEEE}
}

@inproceedings{
shwartz-ziv2021tabular,
title={Tabular Data: Deep Learning is Not All You Need},
author={Ravid Shwartz-Ziv and Amitai Armon},
booktitle={8th ICML Workshop on Automated Machine Learning (AutoML) },
year={2021},
url={https://openreview.net/forum?id=vdgtepS1pV}
}

@inproceedings{
zeng2024tabflex,
title={TabFlex: Scaling Tabular Learning to Millions with Linear Attention},
author={Yuchen Zeng and Wonjun Kang and Andreas C Mueller},
booktitle={NeurIPS 2024 Third Table Representation Learning Workshop},
year={2024},
url={https://openreview.net/forum?id=f8aganC0tN}
}

@article{zabergja2026end,
  title={End-to-End Compression for Tabular Foundation Models},
  author={Zab{\"e}rgja, Guri and Kamel, Rafiq and Kadra, Arlind and Frey, Christian MM and Grabocka, Josif},
  journal={arXiv preprint arXiv:2602.05649},
  year={2026}
}

@ARTICLE{borisov2024,
  author={Borisov, Vadim and Leemann, Tobias and Seßler, Kathrin and Haug, Johannes and Pawelczyk, Martin and Kasneci, Gjergji},
  journal={IEEE Transactions on Neural Networks and Learning Systems}, 
  title={Deep Neural Networks and Tabular Data: A Survey}, 
  year={2024},
  volume={35},
  number={6},
  pages={7499-7519},
  doi={10.1109/TNNLS.2022.3229161}}

@inproceedings{fu2024serverlessllm,
  title={$\{$ServerlessLLM$\}$:$\{$Low-Latency$\}$ serverless inference for large language models},
  author={Fu, Yao and Xue, Leyang and Huang, Yeqi and Brabete, Andrei-Octavian and Ustiugov, Dmitrii and Patel, Yuvraj and Mai, Luo},
  booktitle={18th USENIX Symposium on Operating Systems Design and Implementation (OSDI 24)},
  pages={135--153},
  year={2024}
}

@article{Menghani2023,
author = {Menghani, Gaurav},
title = {Efficient Deep Learning: A Survey on Making Deep Learning Models Smaller, Faster, and Better},
year = {2023},
issue_date = {December 2023},
publisher = {Association for Computing Machinery},
address = {New York, NY, USA},
volume = {55},
number = {12},
issn = {0360-0300},
url = {https://doi.org/10.1145/3578938},
doi = {10.1145/3578938},
journal = {ACM Comput. Surv.},
month = mar,
articleno = {259},
numpages = {37}
}

@Inbook{Jeyaraman2025,
author="Jeyaraman, Brindha Priyadarshini",
title="Deployment Strategies for LLMs",
bookTitle="Large Language Models Ops for Finance: A Practical Guide to Infrastructure, Implementation, and Innovation",
year="2025",
publisher="Apress",
address="Berkeley, CA",
pages="103--134",
isbn="979-8-8688-1700-7",
doi="10.1007/979-8-8688-1700-7_4",
url="https://doi.org/10.1007/979-8-8688-1700-7_4"
}

@inproceedings{liutabpfn,
  title={TabPFN Unleashed: A Scalable and Effective Solution to Tabular Classification Problems},
  author={Liu, Siyang and Ye, Han-Jia},
  booktitle={Forty-second International Conference on Machine Learning},
  year={2025}
}

@misc{seward_bzip2_1996,
	title = {Bzip2 and {Libbzip2}},
	url = {https://sourceware.org/bzip2/},
	author = {Seward, Julian},
	year = {1996},
	note = {Publication Title: bzip2 : Home},
}

@inproceedings{
hu2022lora,
title={Lo{RA}: Low-Rank Adaptation of Large Language Models},
author={Edward J Hu and yelong shen and Phillip Wallis and Zeyuan Allen-Zhu and Yuanzhi Li and Shean Wang and Lu Wang and Weizhu Chen},
booktitle={International Conference on Learning Representations},
year={2022},
url={https://openreview.net/forum?id=nZeVKeeFYf9}
}

@article{hinton2015distilling,
  title={Distilling the knowledge in a neural network},
  author={Hinton, Geoffrey and Vinyals, Oriol and Dean, Jeff},
  journal={arXiv preprint arXiv:1503.02531},
  year={2015}
}
\bibliographystyle{icml2026}

\newpage
\appendix
\onecolumn
\section{Additional Details} 

\subsection{Datasets}
\label{sec:datasets} 

All datasets were evaluated by standardised train:test split as provided by Open ML dataset tasks. Details of the datasets used are provided below:

\begin{center}
\setlength{\tabcolsep}{3pt}

\begin{tabular}{lrrrl}
\toprule
Dataset & Samples & Features & Classes & Description \\
\midrule
Amazon employee access & 32,769 & 9 & 2 & Amazon employee access prediction \\
anneal & 898 & 38 & 5 & Steel annealing process classification  \\
Bank Customer Churn & 10,000 & 10 & 2 & Bank customer churn prediction \\
Bank marketing & 45211 & 13 & 2 & Bank marketing classification \\
blood-transfusion-service-center & 748 & 4 & 2 & Blood transfusion donor behavior \\
churn & 5,000 & 19 & 2 & Telecom customer churn prediction \\
coil2000 insurance policies & 9,822 & 85 & 2 & Insurance policy purchase prediction \\
credit card clients default & 30,000 & 23 & 2 & \\
credit-g & 1,000 & 20 & 2 & German credit risk assessment  \\
diabetes & 768 & 8 & 2 & Pima Indians diabetes diagnosis \\
E-CommereShippingData & 10,999 & 10 & 2 &  \\
Fitness Club & 1,500 & 6 & 2 & Fitness club membership churn prediction \\
hazelnut-spread-contaminant-detection & 2,400 & 30 & 2 & Food contamination detection \\
heloc & 10,459 & 23 & 2 &  \\
HR Analytics Job Change of Data Scientist & 19,158 & 12 & 2 &  \\
in vehicle coupon recommendation & 12,684 & 24 & 2 &  \\
Is-this-a-good-customer & 1,723 & 13 & 2 & Customer quality classification \\
jm1 & 10,885 & 21 & 2 &  \\
Marketing Campaign & 2,240 & 25 & 2 & Marketing campaign response prediction \\
maternal health risk & 1,014 & 6 & 3 & Maternal health risk classification \\
MIC & 1,699 & 111 & 8 & Text/molecular classification  \\
NATICUSdroid & 7,491 & 86 & 2 & Android malware detection \\
online shoppers intention & 12,330 & 17 & 2 &  \\
polish companies bankruptcy & 5,910 & 64 & 2 & Polish company bankruptcy prediction \\
qsar-biodeg & 1,054 & 41 & 2 & Chemical biodegradability prediction \\
seismic-bumps & 2,584 & 15 & 2 & Mining seismic hazard prediction \\
splice & 3,190 & 60 & 3 & DNA splice junction classification \\
students dropout and academic success & 4,424 & 36 & 3 & Student academic outcome prediction \\
taiwanese bankruptcy prediction & 6,819 & 94 & 2 & Taiwan company bankruptcy prediction \\
website phishing & 1,353 & 9 & 3 & Phishing website detection \\
\bottomrule
\end{tabular}
\end{center}

\clearpage

\subsection{Classical Comparisons}
\paragraph{Baseline Methods}
\label{sec:baselines}
Baseline models were run for comparison using fixed default configurations across multiple independent datasets. The results provided are the mean across all datasets for each model. Chosen parameters are included for reference:

\begin{verbatim}
    XGBoost
        n_estimators: 100
        max_depth: 6
        learning_rate:0.1
        random_state: 42
        n_jobs: -1
        eval_metric: 'logloss'
        objective: 'binary:logistic' (binary) or 'multi:softmax' (multi-class)
        use_label_encoder: False
    
    CatBoost 
        iterations: 200 
        depth: 6 
        learning_rate: 0.1 
        random_state: 42 
        verbose: False 
        thread_counts: -1 
        loss_function: 'Logloss' (binary) or 'Multiclass' (multi-class) 
    
    LightGBM 
        n_estimators: 100
        max_depth: 6
        learning_rate:0.1
        random_state: 42
        n_jobs: -1
        verbose: -1 
        force_col_wise: True 
    
    RandomForest
        n_estimators: 100
        max_depth: 10
        random_state: 42
        n_jobs: -1
        verbose: 0
    
    KNN
        n_neighbors: 5
        n_jobs: -1
\end{verbatim}

\paragraph{Tuning Methods}      
\texttt{RandomisedSearchCV} was used with stratified 3-fold cross validation (20 random hyperparameter configurations per model) using the training dataset and reporting the optimized model performance on the test data. Multiple parameter grid search spaces were used covering regularization, tree depth, learning rates, and other model-specific parameters. Details of the parameter search grids are included below:
\begin{verbatim}
    XGBoost (9 parameters) 
        max_depth: [3, 5, 7, 10]
        learning_rate: [0.01, 0.05, 0.1, 0.2] 
        n_estimators: [50, 100, 200, 300] 
        min_child_weight: [1, 3, 5]
        subsample: [0.6, 0.8, 1.0] 
        colsample_bytree: [0.6, 0.8, 1.0] 
        gamma: [0. 0.1, 0.2]
        reg_alpha: [0, 0.1, 0.5, 1.0]
        reg_lambda: [0.5, 1.0, 2.0]

    CatBoost (7 parameters)
        iterations: [50, 100, 200, 300]
        depth: [4, 6, 8, 10]
        learning_rate: [0.01, 0.05, 0.1, 0.2]
        l2_leaf_reg: [1, 3, 5, 7]
        border_count: [32, 64, 128]
        bagging_temperature: [0, 0.5, 1.0]
        random_strength: [0.5, 1.0, 2.0]

    LightGBM (9 parameters)
        num_leaves: [15, 31, 63, 127]
        max_depth: [3, 5, 7, 10, -1]
        learning_rate: [0.01, 0.05, 0.1, 0.2]
        n_estimators: [50, 100, 200, 300]
        min_child_samples: [10, 20, 30, None]
        subsample: [0.6, 0.8, 1.0]
        colsample_bytree: [0.6, 0.8, 1.0]
        reg_alpha: [0, 0.1, 0.5, 1.0]
        reg_lambda: [0, 0.1, 0.5, 1.0]

    RandomForest (7 parameters)
        n_estimators: [50, 100, 200, 300]
        max_depth: [5, 10, 20, 30, None]
        min_samples_split: [2, 5, 10]
        min_samples_leaf: [1, 2, 4]
        max_features: ['sqrt', 'log2', 0.5, 0.7]
        bootstrap: [True, False]
        criterion: ['gini', 'entropy']
        
    KNN (5 parameters)
        n_neighbours: [3, 5, 7, 9, 11, 15, 21]
        weights: ['uniform', 'distance']
        metric: ['euclidean', 'manhattan', 'minkowski']
        p: [1, 2, 3]
        algorithm: ['ball_tree', 'kd_tree', 'brute']
\end{verbatim}

\section{Hardware Details}
Experiments were conducted on an AWS SageMaker \texttt{ml.g4dn.12xlarge}
instance (4$\times$ NVIDIA T4 GPUs with 16~GB VRAM each, 48 vCPUs and
192~GB of shared RAM). Four worker processes were run in parallel; each
worker was pinned to one GPU and a dedicated set of 12 vCPUs to prevent
inter-worker contention. RAM was shared across workers rather than
strictly partitioned, as physical memory cannot be isolated in the same
way as compute resources. Per-process peak resident set size (RSS)
during inference is reported in
Table~\ref{tab:inference_memory_speed}.

\clearpage

\section{Ablations}

\subsection{K-means N-bit ablation study}

\begin{table*}[h!]
\centering
\caption{K-means N-bit ablation study}
\label{tab:kmeans_nbit_ablation}
\begin{tabular}{llccccc}
\hline
\textbf{Model} & \textbf{Bit Width} & \textbf{Accuracy} & \textbf{ROC-AUC}  & \textbf{Model Size (MB)} & \textbf{Compression Ratio} \\
\hline

\multirow{4}{*}{TabPFN-v2.5}
& Baseline FP32 & 0.865 & 0.910  & 40.95 & 1.00x \\
& 1-bit & 0.523 & 0.535  & 1.52 & 26.94x \\
& 2-bit & 0.751 & 0.863  & 2.80 & 14.63x \\
& 4-bit & 0.864 & 0.906  & 5.36 & 7.64x \\
\hline

\multirow{4}{*}{TabPFN-v2.6}
& Baseline FP32 & 0.864 & 0.906  & 41.05 & 1.00x \\
& 1-bit & 0.540 & 0.558  & 1.59 & 25.82x \\
& 2-bit & 0.824 & 0.875  & 2.87 & 14.30x \\
& 4-bit & 0.861 & 0.905  & 5.43 & 7.56x \\
\hline

\end{tabular}
\vspace{0.5em}
\small

\textit{Note: Results computed on 5 representative datasets and may differ slightly from 30-dataset averages in Table 1 due to dataset sampling.}

\end{table*}

4-bit quantization achieves the optimal trade-off between compression and performance, recovering over 99\% of baseline accuracy at 7.64× compression. 2-bit quantization offers a viable alternative at 14-15× compression with $\sim$90\% accuracy recovery. 1-bit quantization is impractical for production use due to severe accuracy degradation ($\sim$40\%).

\subsection{Performance stability across datasets}

\begin{table*}[h!]
\centering
\caption{Performance stability study across 30 classification datasets}
\label{tab:stability}
\begin{tabular}{llcccc}
\hline

\textbf{Model} & \textbf{Method} & \textbf{Accuracy} & \textbf{ROC-AUC} & \textbf{Balanced Acc} \\
\hline

\multirow{4}{*}{TabPFN-v2.5}
& awq INT4      & 0.869$\pm$0.087 & 0.866$\pm$0.090 & 0.712$\pm$0.175 \\
& baseline FP32 & 0.872$\pm$0.086 & 0.869$\pm$0.089 & 0.720$\pm$0.172 \\
& gptq INT4     & 0.869$\pm$0.087 & 0.866$\pm$0.090 & 0.710$\pm$0.174 \\
& kmeans INT4   & 0.871$\pm$0.085 & 0.867$\pm$0.090 & 0.718$\pm$0.171 \\
\hline

\multirow{4}{*}{TabPFN-v2.6}
& awq INT4      & 0.868$\pm$0.089 & 0.866$\pm$0.091 & 0.714$\pm$0.170 \\
& baseline FP32 & 0.872$\pm$0.086 & 0.870$\pm$0.090 & 0.722$\pm$0.170 \\
& gptq INT4     & 0.867$\pm$0.088 & 0.865$\pm$0.091 & 0.714$\pm$0.171 \\
& kmeans INT4   & 0.869$\pm$0.087 & 0.868$\pm$0.090 & 0.721$\pm$0.167 \\
\hline

\end{tabular}

\small
\textit{Note: Results are averaged across 30 datasets. Standard deviation represents performance variation across datasets.}

\end{table*}

Standard deviation across datasets remains consistent between quantized and baseline variants, indicating that quantization does not introduce additional performance variance across different data distributions.









\clearpage

\subsection{Inference memory and speeds}

\begin{table*}[h!]
\centering
\caption{Inference memory and speed averaged across 30 OpenML datasets.}
\label{tab:inference_memory_speed}
\begin{tabular}{llcccccc}
\hline
\textbf{Model} & \textbf{Method} & \textbf{Peak RSS (MB)} & \textbf{Inference Mem* (MB)} & \textbf{$\Delta$ vs Baseline} & \textbf{Speedup vs FP32} \\
\hline


\multirow{4}{*}{TabPFN-v2.5}
& FP32        & 5,893 & 4,737 & -             & 1.00x \\
& GPTQ INT4   & 5,953 & 4,798 & +61 (+1.3\%)   & 1.10x \\
& KMeans INT4 & 5,867 & 4,712 & -25 (-0.5\%)   & 0.99x \\
& AWQ INT4    & 5,923 & 4,767 & +30 (+0.6\%)   & 1.05x \\
\hline

\multirow{4}{*}{TabPFN-v2.6}
& FP32        & 5,717 & 4,561 & -             & 1.00x \\
& GPTQ INT4   & 5,791 & 4,634 & +73 (+1.6\%)   & 1.00x \\
& KMeans INT4 & 5,727 & 4,571 & +10 (+0.2\%)   & 0.91x \\
& AWQ INT4    & 5,775 & 4,619 & +57 (+1.3\%)   & 1.00x \\
\hline

\end{tabular}

\vspace{0.5em}
\small
\textit{*Inference Mem = Peak RSS - Baseline RSS (process overhead removed).}

\end{table*}

Table~\ref{tab:inference_memory_speed} presents an analysis of inference efficiency. Our primary contribution focuses on weight compression to address deployment bottlenecks (model distribution and storage), achieving up to 7× compression. As an additional analysis, we measure inference memory, which remains largely unchanged ($<2$\% variation, $\sim$4.7~GB), as expected since memory usage is dominated by activations. This observation motivates future work on activation quantization. Inference speed remains stable (±10\%), indicating no performance degradation from weight quantization.



\end{document}